\documentclass{article} 
\usepackage[preprint]{icml2026}

\usepackage{amsmath,amsfonts,bm}

\def\eqref#1{equation~\ref{#1}}

\def\1{\bm{1}}

\DeclareMathAlphabet{\mathsfit}{\encodingdefault}{\sfdefault}{m}{sl}
\SetMathAlphabet{\mathsfit}{bold}{\encodingdefault}{\sfdefault}{bx}{n}

\usepackage{xspace}
\usepackage{amsthm}
\usepackage{subcaption}
\usepackage{hyperref}
\hypersetup{
    colorlinks,
    urlcolor=Cerulean,
    linkcolor=NavyBlue,
    citecolor=PineGreen,
}
\usepackage{url}
\usepackage{enumitem}
\usepackage{graphicx}
\usepackage{afterpage}
\usepackage[dvipsnames]{xcolor}
\usepackage{cleveref}
\newcommand{\mult}{\texttt{mult}\xspace}
\newcommand{\sudoku}{\texttt{sudoku}\xspace}
\newcommand{\qwen}{\texttt{Qwen2.5}\xspace}
\newcommand{\gemma}{\texttt{gemma-3}\xspace}
\newcommand{\llama}{\texttt{Llama-3.2}\xspace}

\newcommand{\Qwen}{\texttt{Qwen2.5-Math-1.5B-Instruct}\xspace}
\newcommand{\Gemma}{\texttt{gemma-3-1B-it}\xspace}
\newcommand{\Llama}{\texttt{Llama-3.2-1B-Instruct}\xspace}

\newtheorem{theorem}{Theorem}[section]
\newtheorem*{theorem*}{Theorem}

\newtheorem{definition}{Definition}[section]
\newtheorem{remark}[theorem]{Remark}

\crefname{definition}{Definition}{Definitions}
\Crefname{definition}{Definition}{Definitions}
\Crefname{figure}{Figure}{Figures}
\crefname{figure}{Figure}{Figures}

\icmltitlerunning{Synthetic Error Injection Fails to Elicit Self-Correction In Language Models}

\begin{document}

\twocolumn[
  \icmltitle{Synthetic Error Injection Fails to Elicit Self-Correction In Language Models}



  \icmlsetsymbol{equal}{*}

  \begin{icmlauthorlist}
    \icmlauthor{David X. Wu}{equal,ucb}
    \icmlauthor{Shreyas Kapur}{equal,ucb}
    \icmlauthor{Anant Sahai}{ucb}
    \icmlauthor{Stuart Russell}{ucb}
  \end{icmlauthorlist}

  \icmlaffiliation{ucb}{Department of EECS, UC Berkeley, Berkeley, CA, USA}

  \icmlcorrespondingauthor{David X. Wu}{david\_wu@berkeley.edu}
  \icmlcorrespondingauthor{Shreyas Kapur}{srkp@berkeley.edu}

  \icmlkeywords{Supervised Fine-Tuning, Reinforcement Learning, Self-Correction, LLMs, Reasoning}

  \vskip 0.3in
]



\printAffiliationsAndNotice{\icmlEqualContribution. Author ordering determined by best-of-three Mario Kart in Sunshine Airport.}

\begin{abstract}
Reinforcement learning has become the dominant paradigm for eliciting reasoning and self-correction capabilities in large language models, but its computational expense motivates exploration of alternatives. Inspired by techniques from autonomous driving and robotics, we investigate whether supervised learning with synthetic error injection can induce self-correction abilities in language models. Our approach inserts artificial errors into reasoning chains, masks them, and supervises the model to recognize and correct these mistakes. Despite the intuitive appeal of this method, we find that it fails to significantly improve performance even on simple synthetic tasks across multiple models. Moreover, even when the model catches its own error, it often parrots the original mistake. We find that the distribution shift of synthetic errors to on-policy errors significantly degrades the error-correction capabilities of the fine-tuned model, even with good synthetic coverage of on-policy errors. 
Our results help explain why on-policy reinforcement learning methods have proven uniquely effective for eliciting self-correction. 
\end{abstract}

\section{Introduction}

\begin{figure*}[ht]
    \centering
    \includegraphics[width=\linewidth]{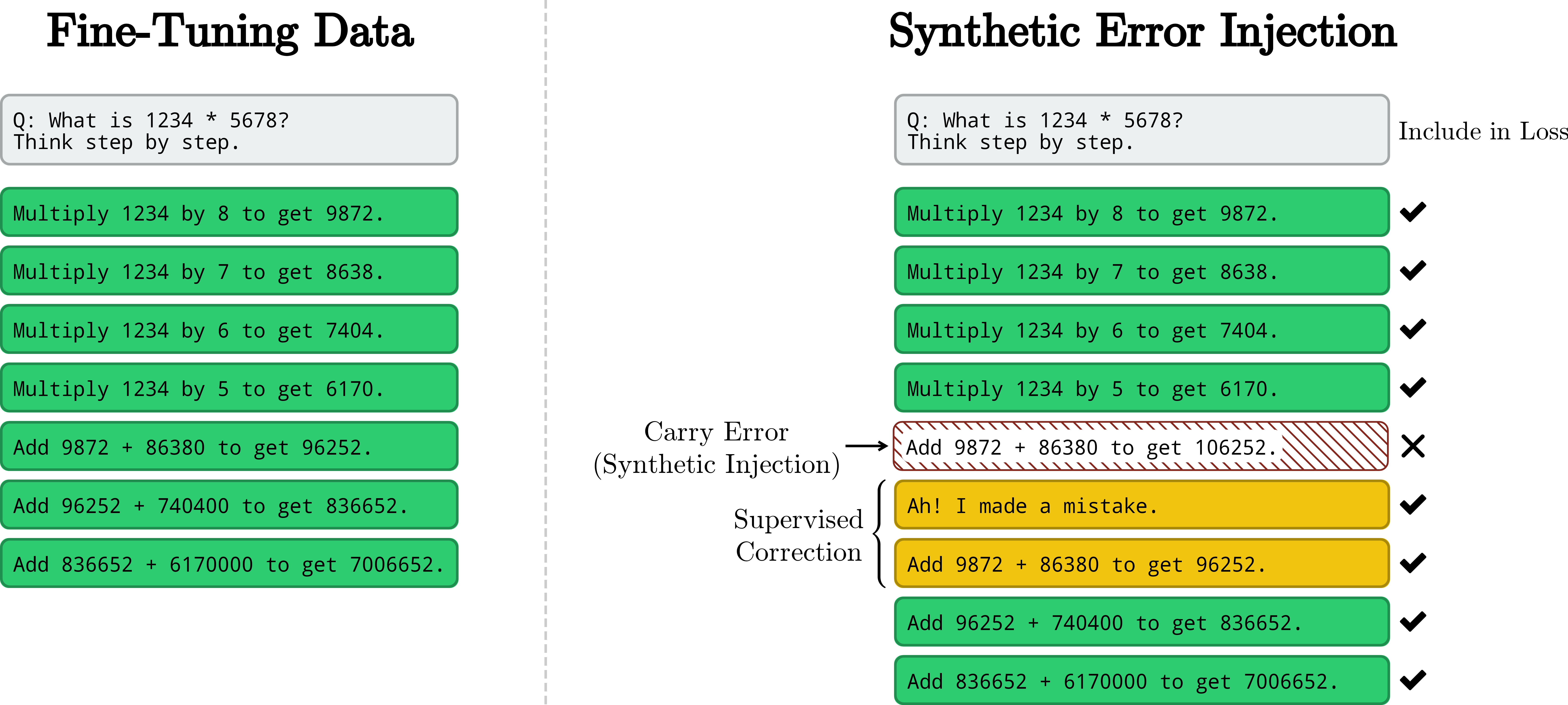}
    \caption{An illustration of the synthetic error injection process. The (Left) panel shows a correct Chain-of-Thought (CoT) trace used for standard fine-tuning (FT). 
    The (Right) panel details the Error Injection Fine-Tuning (EIFT) methodology, where a correct step from a golden CoT is replaced by a three-part sequence: (1) a synthetically injected erroneous step (e.g., a ``Carry Error''), (2) an explicit error recognition step (``Ah! I made a mistake.''), and (3) the original correct step as a supervised correction. As indicated by the ``Include in Loss?'' column, the loss is computed for the recognition and correction steps but masked for the synthetically injected error.}
    \label{fig:hero}
\end{figure*}

The current paradigm for eliciting reasoning in large language models is reinforcement learning (RL) \citep{grpo}. These methods have driven substantial gains in mathematics and coding benchmarks across both closed and open-source models \citep{o1,gemini,kimik2,tulu3}. Such systems often generate an extended chain of thought (CoT) before producing a final answer; within these CoTs, we observe emergent self-reflection and error correction, as the model outlines a solution strategy and then identifies and revises mistakes.

Is it possible to obtain these self-reflection abilities without resorting to reinforcement learning, which is often computationally expensive? The robotics literature offers an illustrative analog. \citet{comma1} trained a self-driving policy purely via imitation learning \citep{comma2}, where the ground-truth trajectory is always correct. When deployed on a real car, the policy gradually drifted until it straddled the lane boundary.

This behavior is a classic instance of distribution shift \citep{distshift}: as errors in real-world dynamics accumulate, a model trained only on ``good'' observations has little guidance on how to recover to the center of the lane. \citet{comma1} mitigated this by artificially and randomly perturbing the camera pose during training using an explicit world model, then supervising the policy to steer back into the lane.

In this paper, we adopt a similar idea for reasoning language models (\Cref{fig:hero}). We use a purely supervised approach: insert synthetic errors into the CoT, mask them, and add a supervised correction step so the model learns to recognize and recover from its own mistakes.

Surprisingly, even on toy tasks, the method fails to yield significant performance improvements across all tested models. We explore two potential hypotheses:
\begin{enumerate}[label=(\roman*)]
    \item We investigate the ``solver-verifier gap''---the  difference in difficulty of generating and checking a step---but find that the method's failure persists even when checking is significantly easier.
    \item We propose that the true issue is a distribution mismatch. While our synthetic error distribution successfully covers the overall distribution of real errors, the model only learns to reliably correct synthetic errors. It fails to generalize this error-correction capability to the distribution of context-dependent errors that the base model itself is prone to making.
\end{enumerate}

Our contributions in this paper are therefore twofold. First, we formally test the hypothesis that supervised error injection can elicit self-correction in LLMs, and we demonstrate its empirical failure. Second, our analysis pinpoints the reason for this failure: a critical mismatch between synthetic error distributions and the model's own innate error modes. This finding suggests that successful self-correction requires more precisely matching error distributions: it must target the specific, context-dependent failures a model is prone to, suggesting why on-policy methods like RL have proven uniquely effective.

\section{Related work}
Chain-of-thought (CoT) reasoning \citep{cot} is a popular approach for imparting reasoning abilities to language models. Recently, reinforcement learning (RL) \citep{o1, o3}---especially stabilization techniques such as GRPO \citep{grpo}---have emerged as the dominant paradigm for teaching models to reason and to self-correct. There is growing interest in this direction, including work such as \citet{kumar2024training}, which uses RL to train models to recognize and correct their own errors.

There has also been substantial effort to improve reasoning and self-correction abilities \textit{without} RL \citep{shinn2023reflexion}, given its computational cost. Approaches include using one LLM as a feedback model for another \citep{madaan2023self, schick2023toolformer}, generate-then-rank strategies \citep{he2022rethinking, weng2023large}, and feedback engines that guide autoregressive decoding \citep{yang2022generating, xie2023self}. \citet{s1}, for instance, proposes a simple technique: append the token “Wait” multiple times when the model attempts to stop, thereby lengthening the reasoning chain and improving performance—without any reinforcement learning.

\citet{kumar2024training} also shows that offline–online distribution mismatch can yield poor on-policy self-correction. We examine this failure mode on even simpler tasks than \texttt{MATH} \citep{math}, specifically small multiplication and Sudoku problems. By constructing a covering set of plausible modeling errors, we show that even such coverage is insufficient, strengthening their result.

Synthetic error injection is widely used in robotics \citep{comma1, comma2}. The work most closely related to ours is \citet{yang2025step}, which trains a language model to play Countdown and to backtrack in a tree-search harness using only supervised data and synthetic error injection. However, their method primarily distills the DFS procedure used to solve Countdown, rather than teaching the model to correct its own errors, and requires a costly sampling harness to generate solutions. In contrast, we teach the model to backtrack fully in a single linear CoT.

\section{Experimental setup}
We trained our models on a cluster of A100 and A6000 GPUs. 
For each model in $\{\Qwen,\allowbreak\Gemma,\allowbreak \Llama\}$ and task in $\{\mult, \sudoku\}$, we trained two versions of the model which differed only in whether the data contained error-injection CoTs.
To solidify the terminology, we define the model variants as follows.
\begin{definition}[FT model]
    We refer to the model trained only on clean CoTs as the \emph{FT model} (Fine Tuned).
\end{definition}
\begin{definition}[EIFT model]
    We refer to the model trained on error injected CoTs as the \emph{EIFT model} (Error Injection Fine Tuned). 
\end{definition}
The details of the EIFT training methodology and data mix can be found in \Cref{sec:eift}; see \Cref{app:training} for training hyperparameters.

\subsection{Error injection}\label{sec:eift}

\begin{figure*}[ht]
    \centering
    \includegraphics[width=\linewidth]{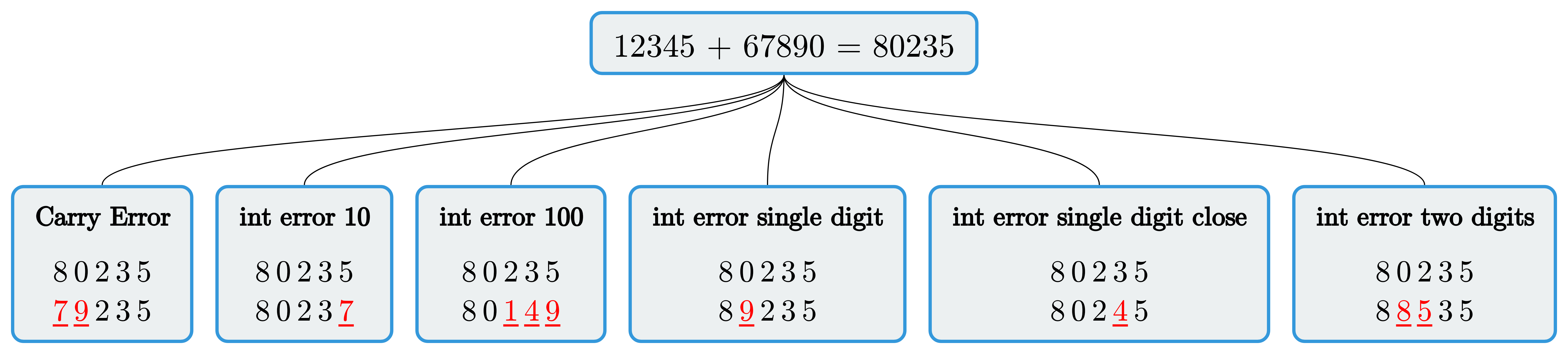}
    \caption{Types of errors we inject for the multiplication task. We take a correct step, depicted here as the addition step at the top of the figure. We then apply one of six types of errors with various probabilities; see \Cref{tab:error_types} for more details. The red underlined digits in each box depict the result of applying the corresponding error type.}
    \label{fig:error_types}
\end{figure*}

For the EIFT model, we used a data mix where $80\%$ of the CoTs were completely correct, and $20\%$ had 1 to 4 synthetic errors injected, where the number and locations of errors were sampled uniformly at random (and without replacement for locations). 
We found that using more complex sampling schemes for the locations of errors did not improve the error correction abilities. 
Since the notion of a synthetic error is task-specific, each task is required to implement the distribution of synthetic errors that it attempts to model for the EIFT training.
To determine this distribution, we trained FT models for various base models and examined the types of errors they tended to make; see the sections below for details.
We validated this overall methodology in \Cref{subsec:error-dist}.

To inject a single error, we replace a golden step $s_i$ at position $i$ in the golden CoT with three steps according to the following procedure:
\begin{enumerate}
    \item The erroneous step $s_i'$, which can easily be verified to be incorrect since we have access to the golden step $s_i$.
    \item The error recognition step, consisting of the tokens \textsc{Ah! I made a mistake}.
    \item Finally, the error correction step, which is just the original golden step $s_i$. 
\end{enumerate}  
For EIFT training, we trained the model to predict the error recognition and error correction steps, and used loss masking to prevent the model from directly predicting the incorrect step $s_i'$. 
However, we did not modify the attention mask, as we wanted the model to learn an internal representation of what errors look like.
We also ablated the token-level weight on the recognition and correction steps (\Cref{fig:correction_weight}), but did not find any significant improvement in recognition or correction rates. 
Thus, for simplicity we did not use any special loss weighting beyond the loss masking.

\subsection{Multiplication}\label{subsec:mult-task}
Our multiplication task consisted of uniformly random 4 digit multiplication problems. 
We found that 4 digit problems were the sweet spot for task difficulty: at this complexity, the base models had low accuracy, and the FT model was significantly more accurate than the base model. 
\Cref{fig:hero} illustrates what the different types of CoTs look like for multiplication; see \Cref{app:examples} for more examples.

\paragraph{Error model.} To model the errors for multiplication, we trained the FT model for \qwen and a held out \texttt{gemma2-2b-it}. We then  manually inspected the types of errors these models tended to make. 
Based on this study, we came up with the following broad error categories:
\begin{itemize}
    \item \texttt{carry\_error}: in an addition step, consider all locations where the column (including carry digits) sums to $9$ or $10$. Pick a random location and swap the column sum to the other value. Swapping $9$ to $10$ means incorrectly adding a carry, and swapping $10$ to $9$ means incorrectly omitting a carry.
    \item \texttt{\_int\_error\_10}: add a random integer in $[-10, 10]$ to the answer.
    \item \texttt{\_int\_error\_100}: add a random integer in $[-100, 100]$ to the answer.
    \item \texttt{\_int\_error\_single\_digit}: replace a randomly chosen single digit of the answer with a different random digit from $0$ to $9$.
    \item \texttt{\_int\_error\_single\_digit\_close}: replace a randomly chosen single digit of the answer with a different random digit within $2$ of it (respecting edge cases to keep the number syntactically valid after error injection).
    \item \texttt{\_int\_error\_two\_digits}: replace a random substring of two contiguous digits with two randomly chosen digits. 
\end{itemize}
For the specific frequencies and a visual depiction of these error types, see
\Cref{tab:error_types,fig:error_types}.

\begin{table*}[!ht]
    \centering
    \begin{tabular}{lcc} \hline
        Error type & $\Pr[\text{error}\mid\text{addition}]$ & $\Pr[\text{error}\mid\text{non-addition}]$ \\ \hline
        \texttt{carry\_error}                            & 0.5   & 0    \\
        \texttt{\_int\_error\_10}                       & 0.025 & 0.05 \\
        \texttt{\_int\_error\_100}                      & 0.025 & 0.05 \\
        \texttt{\_int\_error\_single\_digit}           & 0.125 & 0.25 \\
        \texttt{\_int\_error\_single\_digit\_close}     & 0.125 & 0.25 \\
        \texttt{\_int\_error\_two\_digits}              & 0.20  & 0.40 \\ \hline
    \end{tabular}
    \caption{Error probabilities for different step types. For addition steps, with probability $0.5$ we inject an addition-carry error; otherwise we sample from the standard integer error injector (the \texttt{\_int\_error\_*} distribution). For non-addition steps we always use the standard injector.}
    \label{tab:error_types}
\end{table*}
To pick the frequencies of the errors, we hill climbed on the final accuracy for the held out model, \texttt{gemma2-2b-it}.
\subsection{Sudoku}

\begin{figure*}[ht]
    \centering
    \includegraphics[width=\linewidth]{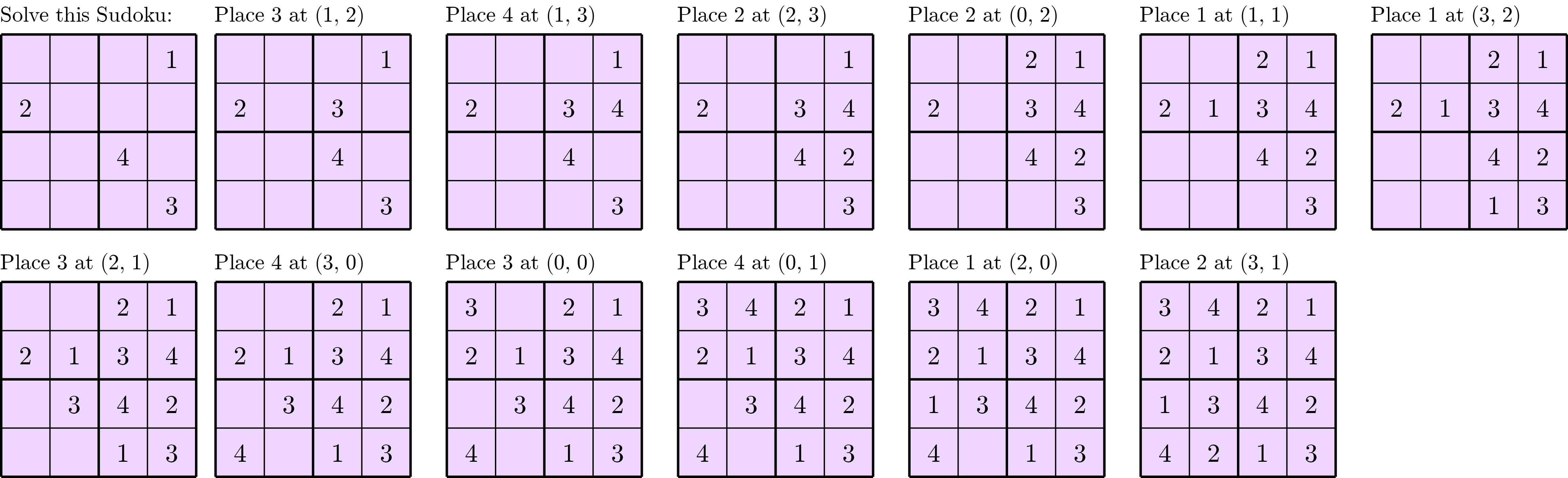}
    \caption{An example of a golden Chain-of-Thought (CoT) for the $4 \times 4$ Sudoku task. The solution trajectory is restricted to "naked single" moves, defined as instances where a specific cell has exactly one possible candidate number remaining. The problem instances are generated by solving a random board and removing numbers such that a naked single solution path remains viable.}
    \label{fig:sudoku-golden}
\end{figure*}

The Sudoku task consisted of $4 \times 4$ Sudoku problems. 
To make the problem more tractable for the models, we restricted to golden solutions that only make ``naked single'' moves.
\begin{definition}[Naked single move]
    For a given board state, a \emph{naked single} is a cell that has exactly one possible candidate number remaining. 
\end{definition}
Correspondingly, we needed to sample Sudoku problems with naked singles solutions. To do so, we first generated problems by using a Sudoku solver with randomness to generate a random valid solution.
We then randomly removed numbers one by one and rejected candidates where there are no naked singles in the new board. 
See \Cref{fig:sudoku-golden} for an example of golden CoTs for sudoku.

\paragraph{Error model.} To model errors, we used an equal split of invalid moves (i.e. moves with invalid coordinates or moves which violate a constraint) and moves that are not a naked single.
We found that only modeling errors that are invalid moves did not impart error correction capabilities in the EIFT model.
\subsection{Evaluation}\label{subsec:eval}
For evaluation, we sampled 1000 problems for each task and evaluated the FT and EIFT models using greedy sampling. 
We also ablated the temperature setting (across $T=0.7, 1.0$), but did not find any significant difference in the accuracy or error correction capabilities; see \Cref{sec:results,fig:temp} for more discussion.

To measure the accuracy of our models, we only graded their final answers. 
Even at modest dataset sizes, the models are very consistent with the exact answer format that we finetuned them on.
Hence, we did not observe any false positives or negatives in our grading.

To measure the error correction capabilities, we follow a more involved procedure. 
Of course, one should not expect the model to correct from arbitrary error distributions.
To narrow the scope, we compare the error correction capabilities where the errors are sampled from the following two error distributions. 
\begin{definition}[Synthetic errors]\label{def:synthetic-error}
    A \emph{synthetic} error is an erroneous step in a CoT generated by our synthetic error injector. 
\end{definition}
\begin{definition}[FT errors]\label{def:ft-error}
    An \emph{FT} error is an erroneous step in a CoT generated by the FT model.
\end{definition}
In other words, we generate synthetic errors by sampling synthetic CoTs and finding errors in them, and we generate FT errors by sampling FT CoTs and finding errors in them. See \Cref{fig:mult_errors_synthetic,fig:mult_errors_ft}
for synthetic and FT errors, respectively, for \qwen on multiplication.

With this in mind, we use the following process to measure error correction for the model. 
\begin{enumerate}
    \item Depending on which error source we are testing, we generate the appropriate type of CoT (synthetic or FT). 
    \item We truncate this CoT at the location of the first error, \emph{including the error itself}. If the CoT does not have any errors, we discard this sample and repeat.
    \item We prompt the model to complete the truncated CoT.
    \item We grade the error recognition and correction for the guaranteed first error step produced by truncation. 
    Namely, if first error step is at step $i$, we check for recognition on step $i+1$ and correction on step $i+2$. 
    Concretely, this entails checking for the tokens \textsc{Ah! I made a mistake} at step $i+1$ and checking that the step $i+2$ is correct \emph{as defined by our task}.
\end{enumerate}
A couple points are in order about the above procedure.
First, an important nuance to clarify is the exact notion of an incorrect step.
To be conservative, we only check for errors that we have modeled.
For multiplication, this means we only check steps where an atomic add or multiply is being performed incorrectly; for Sudoku, this means we check move placement steps and whether they are valid naked singles moves. 
In particular, we do not grade whether the model frivolously error corrects, as long as the proposed correction step is still correct.
    
Also, by convention, error corrections are a subset of error recognition---in order to correct an error, it must first declare that it has made a mistake. 
In practice, we never observed the model doing an error correction without first outputting an error recognition step.

\subsection{Error Distributions}\label{subsec:error-dist}

\begin{figure}[!ht]
    \centering
    \captionsetup[subfigure]{labelfont=rm}
    
    \begin{subfigure}[b]{\linewidth}
        \centering
        \includegraphics[width=\linewidth]{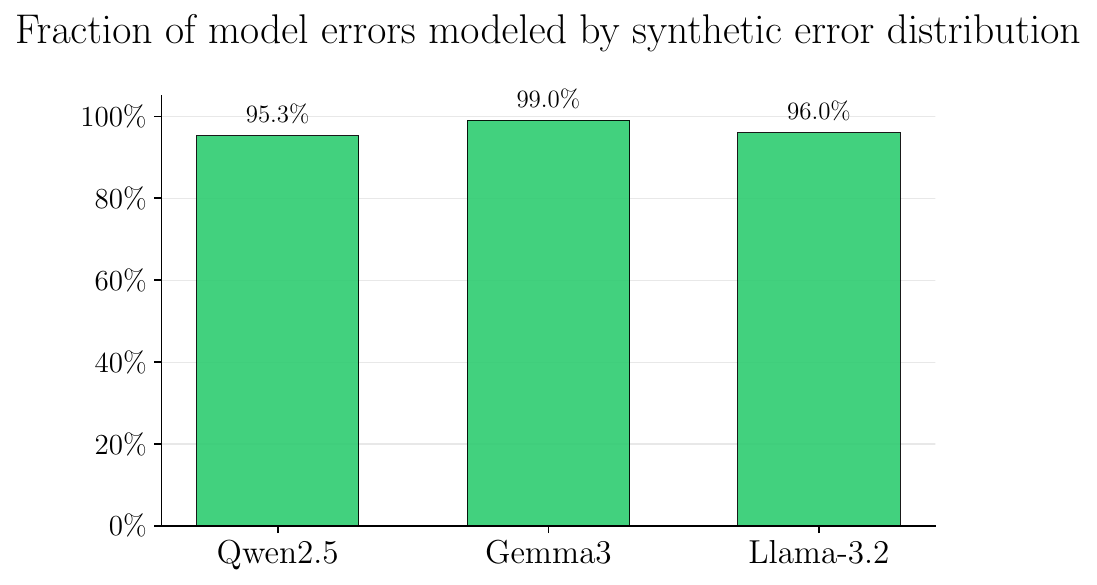}
        \caption{}
        \label{fig:modeled_errors_frac}
    \end{subfigure}
    
    \par\bigskip 
    
    \begin{subfigure}[b]{\linewidth}
        \centering
        \includegraphics[width=\linewidth]{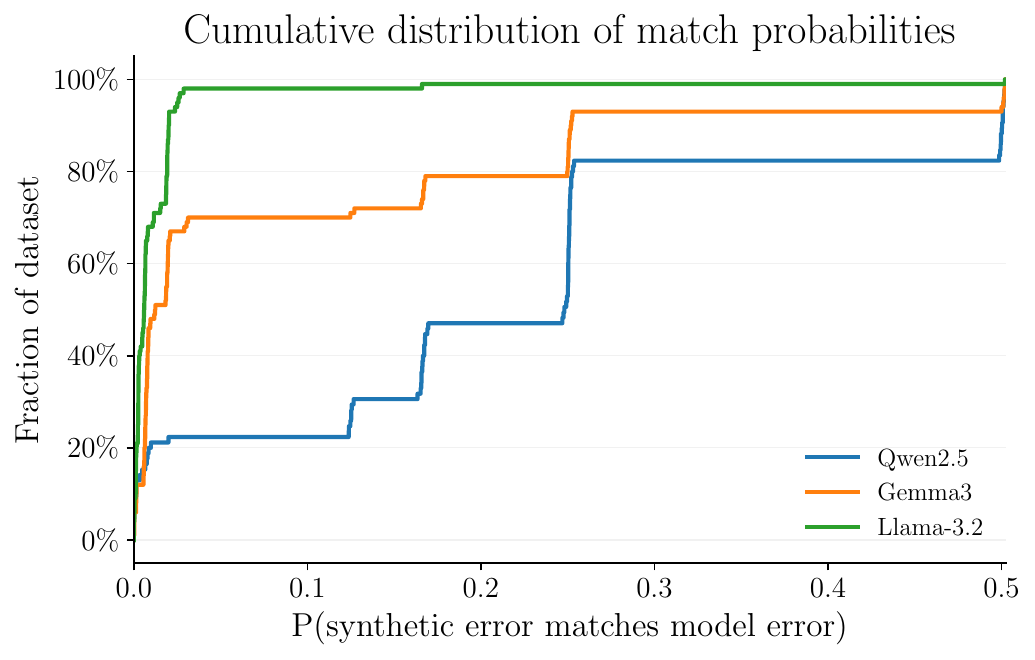} 
        \caption{}
        \label{fig:modeled_errors_cdf}
    \end{subfigure}
    
    \caption{\textbf{Validation of synthetic error distribution alignment with on-policy errors.} 
\textbf{(a) Error Coverage:} The percentage of on-policy model errors that appear at least once within $n=10,000$ samples from our synthetic error injector. We achieve near-perfect coverage ($>95\%$) across all models, indicating the injector correctly identifies the support of the error distribution. 
\textbf{(b) Distributional Alignment:} The cumulative distribution of the probability assigned by the synthetic injector to the \textit{exact} on-policy error step (down and to the right is better). For a given probability threshold $p$ on the $x$-axis, the $y$-axis represents what fraction of the 100 errors could be exactly matched with probability at most $p$. The probability mass placed on exact matches (particularly for \qwen \ and \gemma) indicates that the synthetic distribution closely approximates the frequency of natural model errors.}
    \label{fig:modeled_errors}
\end{figure}

While we qualitatively designed the injector to mimic observed error types, it is crucial to validate if the synthetic distribution accurately covers the on-policy error distribution.
Hence, we sought to quantify the alignment between the synthetic errors used during training and the natural, on-policy errors made by the models. 

We measured this alignment by estimating the probability of \emph{exact} token-level match, i.e. how often the synthetic error injector exactly replicates an FT error.
First, we collected 100 reasoning traces from the FT model containing calculation errors on the multiplication task. 
For each erroneous instance, we truncated the trace immediately prior to the first incorrect step and sampled $n = 10,000$ candidate erroneous steps from our synthetic error injector. 
In comparison, during finetuning we trained the model on approximately $100,000$ incorrect steps. 

\Cref{fig:modeled_errors_frac} illustrates the error coverage: the percentage of on-policy errors that appeared at least once within the $n$ synthetic samples. We observe near-perfect coverage across all models: 95.3\% for \qwen, 99.0\% for \gemma, and 96.0\% for \llama. This indicates that the support of our synthetic distribution encompasses nearly all on-policy error modes.

Further, \Cref{fig:modeled_errors_cdf} demonstrates that our synthetic injector achieves distributional alignment. We plot the empirical cumulative distribution function (CDF) of the probability assigned to the exact on-policy error step among the $100$ reasoning traces. In generative tasks, the space of potential incorrect tokens is vast, making any significant probability mass on an exact string match a strong indicator of alignment. We find that our injector frequently assigns high probabilities to the model's specific natural errors. For instance, it assigns greater than 15\% probability to the exact on-policy errors made by \qwen and \gemma in a large fraction of CoTs (70\% for \qwen and 30\% for \gemma). This implies that during training, the models frequently encounter the \textit{exact} errors they are prone to making, confirming the quality of the synthetic error distribution.

\section{Results}\label{sec:results}
For evaluation, we measured accuracy and error correction/recognition rates using a fixed set of 1000 randomly sampled problems. 
In all the figures below, we also report $1.96 \cdot \text{SE}$ error bars.

\begin{figure*}[!ht]
\begin{center}
\includegraphics[width=\linewidth]{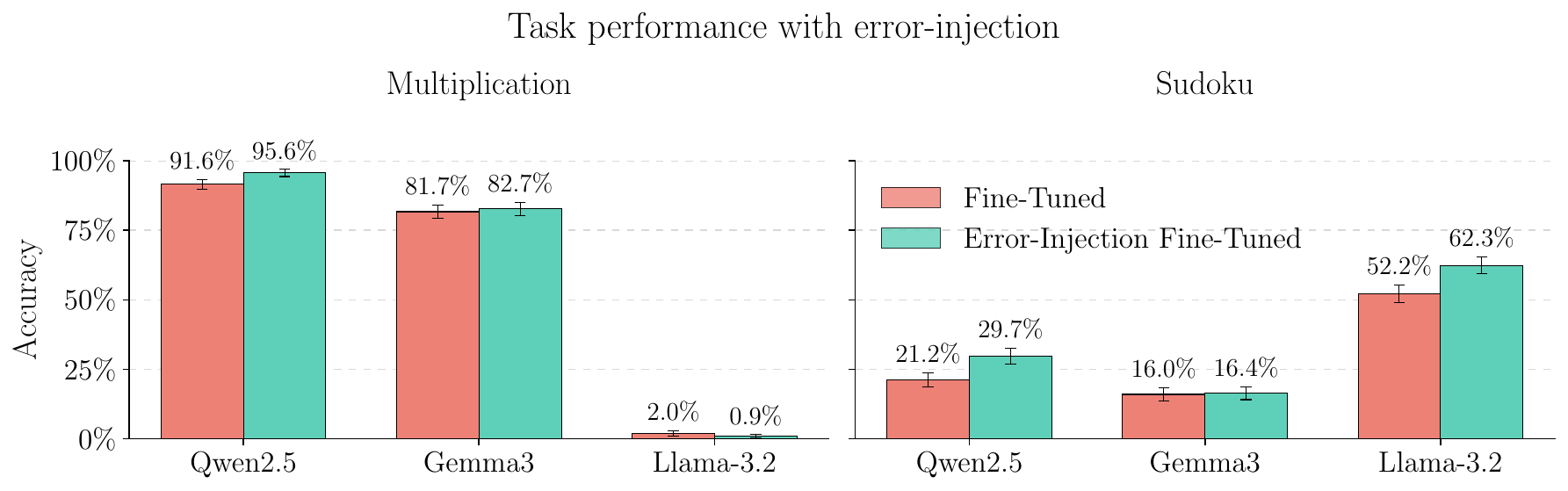}
\end{center}
\caption{A comparison of model performance grouped by base model and task. 
We report the mean accuracy measured on a fixed set of 1000 randomly generated problems, and error bars are 1.96SE.
The two model types are (1) FT (Fine-Tuned, red) models which are trained solely on golden CoTs or (2) EIFT (Error-Injection Fine-Tuned, green) on error-injected CoTs. 
One would expect the EIFT models to have higher accuracy than the FT models. 
However, across the board, we see only modest ($<5\%$) performance gains for multiplication. For Sudoku tasks the performance is boosted by roughly 10\% for Qwen and Llama, but none at all for Gemma.}
\label{fig:accuracy}
\end{figure*}

We hypothesized that the EIFT model would have better performance than the corresponding FT model. 
In \Cref{fig:accuracy} we see that this hypothesis is not strongly supported by the empirical results. 
For multiplication, we see modest to no performance gains from EIFT training.
For Sudoku, we see more nontrivial gains of roughly $10\%$ for \qwen and \llama. However, \gemma shows no performance boost from EIFT training.
The baseline performance of the FT model varied drastically depending on the base model and task. 
For example, \qwen performed quite well (FT performance: 92\%) on multiplication, but was not as strong on Sudoku.
Conversely, \llama had very poor performance on multiplication (FT performance: 2.0\%) but had the best performance on Sudoku (FT performance: 52.2\%). 
See \Cref{fig:mult_errors_eift} for some examples of failure modes of the EIFT model.
We hypothesize that these fluctuations are due to the specific mid/post-training recipes for these models, e.g., since \qwen is a specialized math model.

One plausible explanation for why the performance is not significantly boosted from EIFT training is that the model does not actually reliably correct its own errors.
Thus, we turn to investigating the correction and recognition capabilities of the EIFT model. 

\begin{figure*}[!ht]
\begin{center}
\includegraphics[width=\linewidth]{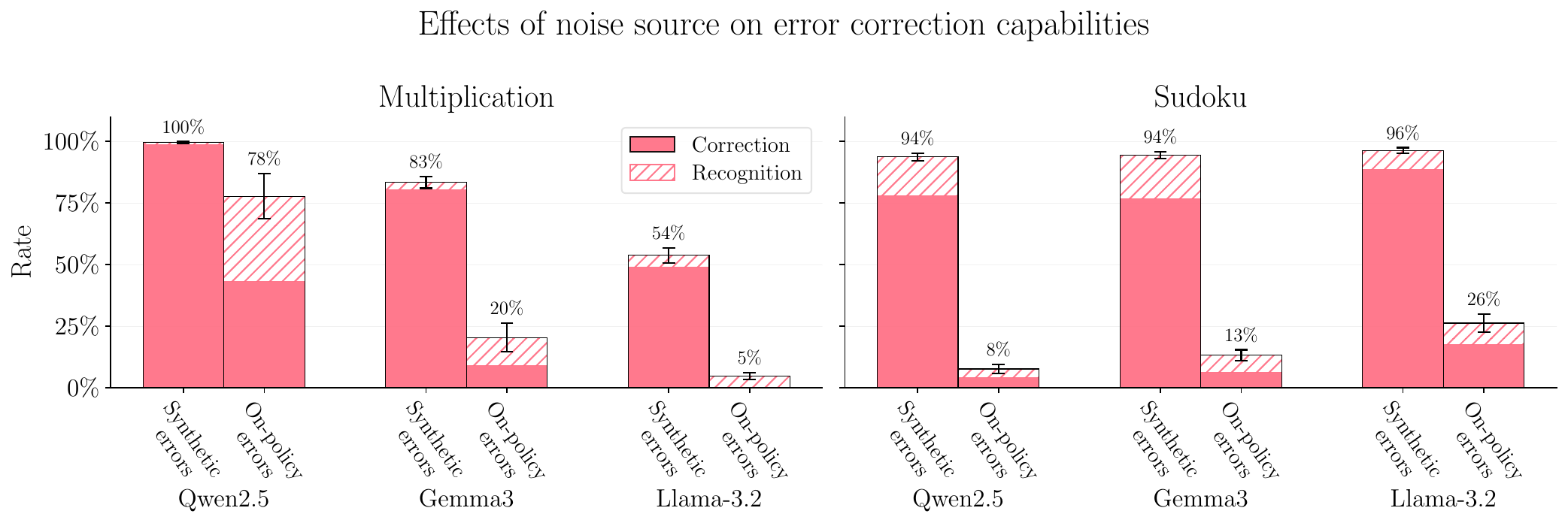}
\end{center}
\caption{Comparison of error correction capabilities for EIFT models based on the error source and task.
We report the mean correction and recognition rates measured on a fixed set of 1000 randomly generated problems, and error bars are 1.96SE.
We compare errors that the EIFT model was trained to correct and recognize (Synthetic errors) versus errors the FT model naturally produces (On-policy errors). 
The recognition rates (hatched) are always larger than the correction rates (solid) as recognition is a superset of correction. 
The error correction capabilities would ideally generalize beyond the exact error distribution that the EIFT model is trained on. 
Aside from Qwen multiplication, we see that both correction and recognition rates drop drastically across all settings and base models. 
In general, there is a significant gap between error correction and error recognition in the on-policy error settings. 
Furthermore, for Sudoku such a gap is present even for synthetic errors. 
}
\label{fig:error}
\end{figure*}

In \Cref{fig:error} we see the EIFT model, despite having  high error recognition rates for the synthetic errors (\Cref{def:synthetic-error}), suffers from a precipitous drop in correction and recognition rates when prompted with on-policy errors generated by the FT model (\Cref{def:ft-error}).
In all cases except for \qwen on multiplication, the FT error recognition rate drastically plummets---e.g., 94\% $\to$ 8\% for \qwen on Sudoku, 83\% $\to$ 20\% for \gemma on multiplication. 
Furthermore, there is a nontrivial gap in correction and recognition rates for FT errors. 
For Sudoku, this gap also exists for synthetic errors, suggesting that error correction is  somewhat nontrivial for this task. 
For Qwen on multiplication, even though the model has better retention in error recognition (100\% $\to$ 78\%), it nevertheless significantly drops in error correction (99\% $\to$ 40\%). Interestingly, in this setting the EIFT model simply repeats the original error in 25\% of the cases where it fails to correct from an FT error.  
See \Cref{fig:mult_errors_eift} for some examples of similar failure modes.
\begin{remark}
To be thorough, we also evaluated the FT models, and as expected the FT models did not possess error correction capabilities.
\end{remark}

\begin{figure*}[h]
\begin{center}
\includegraphics[width=\linewidth]{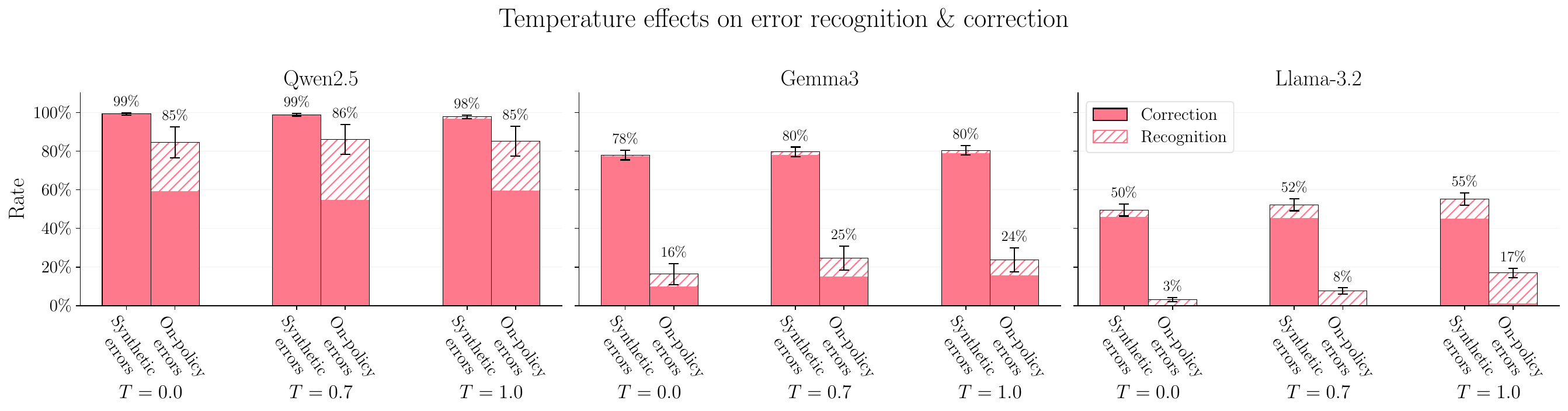}
\end{center}
\caption{Temperature ablation for multiplication evaluation. We compare the correction and recognition rates for the exact same multiplication checkpoints, using different temperature settings: $T \in \{0.0, 0.7, 1.0\}$. We find no significant change in the capabilities across temperature settings for Qwen. For Gemma and Llama, we only see modest gains in performance with positive temperature, but the gap between synthetic and on-policy performance remains large ($>40\%$ absolute difference).}
\label{fig:temp}
\end{figure*}

As mentioned in \Cref{subsec:eval}, we report our model evaluations using greedy sampling. 
We investigate the effect of positive temperature sampling on error correction and recognition rates; one might expect that error correction (and to a lesser extent error recognition) rates would be boosted by positive temperature sampling. 
In \Cref{fig:temp} we test this hypothesis for the multiplication task using $T \in \{0.0, 0.7, 1.0\}$, and observe that in general the correction rates do not significantly change with positive temperature. 
In addition, we observe no significant change in the  recognition rate for \qwen and \gemma. 
Interestingly, \llama benefits the most from positive temperature sampling for error recognition, but remains unable to perform error correction.
We conclude that our findings are relatively robust to the exact sampling temperature.

\section{Discussion}
We have shown that, even in toy synthetic settings,  synthetic error-injected SFT does not recover the performance of online RL, in spite of having access to golden solutions and perfect verifiers. 
At the very least, our results suggest that SFT-based approaches to error correction must essentially match the on-policy distribution, or else inject significantly richer information to condition the model to try different strategies.

\paragraph{Solver-Verifier gap.} 
When we conducted our initial experiments for multiplication, we were surprised that EIFT training did not significantly improve accuracy. 
One intuitive explanation we theorized was that for multiplication, the difficulty of a model verifying or checking an intermediate step is about as hard as doing it correctly in the first place. 
\footnote{Certain heuristics can check intermediate work in arithmetic, such as rules based on modular arithmetic.}

This led to designing the Sudoku task, which does have a more pronounced solver-verifier gap. 
For any particular move, verifying is trivial by just checking the constraints, whereas solving requires searching for the valid next step. 
Indeed, from \Cref{fig:accuracy} we see that EIFT training boosts Sudoku performance more than multiplication performance for the \qwen\, and \llama\,  models. 
It would be interesting to study more systematically the benefits of error injection as a function of the solver-verifier gap.
Some examples of more difficult solver-verifier gaps would be NP-hard search problems (such as general Sudoku where the solution need not be naked singles), but these would likely require training stronger base models than we could reasonably achieve given our compute limitations.

\paragraph{Interpreting error correction.}
We leave open the question of the precise mechanism behind the model's error recognition and correction capabilities.
For example, we found it quite surprising that the model was able to recognize errors significantly more often than it could correct them.
Most notably, we almost always observe that in CoTs where the model recognizes but doesn't correct, it just parrots the original incorrect step; see \Cref{fig:mult_errors_eift} for an example. 
Moreover, this phenomenon was replicated across different temperature settings, suggesting that the model believes the most likely step is still the original (incorrect) step.

There are several different explanations for this type of behavior. 
One possibility is that the model learns two circuits, one for computing the answer to an intermediate step, and the other for verifying said answer. 
However, the model may not have learned how to use the verifier output to alter its circuit to get the correct (or even different) answer.
Under this heuristic picture, it would be unsurprising that the model can rarely error correct.
It would be interesting to confirm whether this hypothesis is accurate and to see if there are any other training recipes that naturally ``diversify'' the strategies the model attempts towards solving the problem.

\paragraph{Error injectors and RL.} 
One natural direction for further exploration is to generalize our results beyond synthetic error injectors. 
We now discuss some obstacles we faced when we tried to move in this direction.

The main issue one needs to overcome is the problem of efficiently generating EIFT datasets. 
Fundamentally, the usability of an EIFT method is bottlenecked by efficient verification, how accurate the error model is, and the availability of a golden solution. 
For example, the most natural error model is to take an on-policy RL approach and only use the model's own errors.
In this setting, the challenge then shifts to constructing an efficient and accurate enough verifier (e.g. a well-trained process reward model). 
Since our method is supposed to avoid doing RL in the first place, we did not pursue this direction further. 

Another possibility is to prompt LLMs to inject more realistic-looking errors. 
The hope here is that cleverly prompted LLMs might provide a scalable way to construct error distributions that are more on-policy.
However, in early experiments, we found that it was somewhat difficult to steer the LLMs to produce errors in a reliable way that was useful for EIFT training. 
For instance, the model would often explicitly acknowledge that it was making a mistake, or it would make implausible or unnatural-looking errors. 

These limitations could potentially be overcome by careful few-shot prompting. 
In practice, the process of constructing and iterating on few-shot prompts starts to look awfully similar to the construction of a synthetic error injector, especially if one wants to tune the frequency of certain types of errors appearing. 
On the other hand, it is possible that in more complicated reasoning scenarios, LLM based error injectors become the only scalable way to model errors.

\subsubsection*{Acknowledgments}
DW was supported by NSF Graduate Research Fellowship DGE-2146752. 
SK was supported in part by the AI2050 program at Schmidt Futures (Grant G-22-63471).

\bibliography{bib}

@article{grpo,
  title={Deepseek-r1: Incentivizing reasoning capability in llms via reinforcement learning},
  author={Guo, Daya and Yang, Dejian and Zhang, Haowei and Song, Junxiao and Zhang, Ruoyu and Xu, Runxin and Zhu, Qihao and Ma, Shirong and Wang, Peiyi and Bi, Xiao and others},
  journal={arXiv preprint arXiv:2501.12948},
  year={2025}
}

@article{cot,
  title={Chain-of-thought prompting elicits reasoning in large language models},
  author={Wei, Jason and Wang, Xuezhi and Schuurmans, Dale and Bosma, Maarten and Xia, Fei and Chi, Ed and Le, Quoc V and Zhou, Denny and others},
  journal={Advances in neural information processing systems},
  volume={35},
  pages={24824--24837},
  year={2022}
}

@article{o1,
  title={{OpenAI o1 system card}},
  author={Jaech, Aaron and Kalai, Adam and Lerer, Adam and Richardson, Adam and El-Kishky, Ahmed and Low, Aiden and Helyar, Alec and Madry, Aleksander and Beutel, Alex and Carney, Alex and others},
  journal={arXiv preprint arXiv:2412.16720},
  year={2024}
}

@inproceedings{comma1,
  title={Learning to drive from a world model},
  author={Goff, Mitchell and Hogan, Greg and Hotz, George and du Parc Locmaria, Armand and Raczy, Kacper and Sch{\"a}fer, Harald and Shihadeh, Adeeb and Zhang, Weixing and Yousfi, Yassine},
  booktitle={Proceedings of the Computer Vision and Pattern Recognition Conference},
  pages={1964--1973},
  year={2025}
}

@misc{comma2,
Author = {Harald Schafer and Eder Santana and Andrew Haden and Riccardo Biasini},
Title = {A Commute in Data: The comma2k19 Dataset},
Year = {2018},
Eprint = {arXiv:1812.05752},
}

@inproceedings{distshift,
  title={A reduction of imitation learning and structured prediction to no-regret online learning},
  author={Ross, St{\'e}phane and Gordon, Geoffrey and Bagnell, Drew},
  booktitle={Proceedings of the fourteenth international conference on artificial intelligence and statistics},
  pages={627--635},
  year={2011},
  organization={JMLR Workshop and Conference Proceedings}
}

@article{gemini,
  title={Gemini 2.5: Pushing the frontier with advanced reasoning, multimodality, long context, and next generation agentic capabilities},
  author={Comanici, Gheorghe and Bieber, Eric and Schaekermann, Mike and Pasupat, Ice and Sachdeva, Noveen and Dhillon, Inderjit and Blistein, Marcel and Ram, Ori and Zhang, Dan and Rosen, Evan and others},
  journal={arXiv preprint arXiv:2507.06261},
  year={2025}
}

@article{o3,
  title={{OpenAI o3 and o4-mini system card}},
  author={OpenAI},
  year={2025}
}

@article{kimik2,
  title={Kimi k2: Open agentic intelligence},
  author={Bai, Yifan and Bao, Yiping and Chen, Guanduo and Chen, Jiahao and Chen, Ningxin and Chen, Ruijue and Chen, Yanru and Chen, Yuankun and Chen, Yutian and others},
  journal={arXiv preprint arXiv:2507.20534},
  year={2025}
}

@article{tulu3,
  title={Tulu 3: Pushing frontiers in open language model post-training},
  author={Lambert, Nathan and Morrison, Jacob and Pyatkin, Valentina and Huang, Shengyi and Ivison, Hamish and Brahman, Faeze and Miranda, Lester James V and Liu, Alisa and Dziri, Nouha and Lyu, Shane and others},
  journal={arXiv preprint arXiv:2411.15124},
  year={2024}
}

@article{yang2025step,
  title={Step back to leap forward: Self-backtracking for boosting reasoning of language models},
  author={Yang, Xiao-Wen and Zhu, Xuan-Yi and Wei, Wen-Da and Zhang, Ding-Chu and Shao, Jie-Jing and Zhou, Zhi and Guo, Lan-Zhe and Li, Yu-Feng},
  journal={arXiv preprint arXiv:2502.04404},
  year={2025}
}

@article{kumar2024training,
  title={Training language models to self-correct via reinforcement learning},
  author={Kumar, Aviral and Zhuang, Vincent and Agarwal, Rishabh and Su, Yi and Co-Reyes, John D and Singh, Avi and Baumli, Kate and Iqbal, Shariq and Bishop, Colton and Roelofs, Rebecca and others},
  journal={arXiv preprint arXiv:2409.12917},
  year={2024}
}

@inproceedings{s1,
  title={s1: Simple test-time scaling},
  author={Muennighoff, Niklas and Yang, Zitong and Shi, Weijia and Li, Xiang Lisa and Fei-Fei, Li and Hajishirzi, Hannaneh and Zettlemoyer, Luke and Liang, Percy and Cand{\`e}s, Emmanuel and Hashimoto, Tatsunori B},
  booktitle={Proceedings of the 2025 Conference on Empirical Methods in Natural Language Processing},
  pages={20286--20332},
  year={2025}
}

@article{madaan2023self,
  title={Self-refine: Iterative refinement with self-feedback},
  author={Madaan, Aman and Tandon, Niket and Gupta, Prakhar and Hallinan, Skyler and Gao, Luyu and Wiegreffe, Sarah and Alon, Uri and Dziri, Nouha and Prabhumoye, Shrimai and Yang, Yiming and others},
  journal={Advances in Neural Information Processing Systems},
  volume={36},
  pages={46534--46594},
  year={2023}
}

@article{schick2023toolformer,
  title={Toolformer: Language models can teach themselves to use tools},
  author={Schick, Timo and Dwivedi-Yu, Jane and Dess{\`\i}, Roberto and Raileanu, Roberta and Lomeli, Maria and Hambro, Eric and Zettlemoyer, Luke and Cancedda, Nicola and Scialom, Thomas},
  journal={Advances in Neural Information Processing Systems},
  volume={36},
  pages={68539--68551},
  year={2023}
}

@inproceedings{weng2023large,
  title={Large language models are better reasoners with self-verification},
  author={Weng, Yixuan and Zhu, Minjun and Xia, Fei and Li, Bin and He, Shizhu and Liu, Shengping and Sun, Bin and Liu, Kang and Zhao, Jun},
  booktitle={Findings of the Association for Computational Linguistics: EMNLP 2023},
  pages={2550--2575},
  year={2023}
}

@article{he2022rethinking,
  title={Rethinking with retrieval: Faithful large language model inference},
  author={He, Hangfeng and Zhang, Hongming and Roth, Dan},
  journal={arXiv preprint arXiv:2301.00303},
  year={2022}
}

@article{yang2022generating,
  title={Generating natural language proofs with verifier-guided search},
  author={Yang, Kaiyu and Deng, Jia and Chen, Danqi},
  journal={arXiv preprint arXiv:2205.12443},
  year={2022}
}

@article{xie2023self,
  title={Self-evaluation guided beam search for reasoning},
  author={Xie, Yuxi and Kawaguchi, Kenji and Zhao, Yiran and Zhao, James Xu and Kan, Min-Yen and He, Junxian and Xie, Michael},
  journal={Advances in Neural Information Processing Systems},
  volume={36},
  pages={41618--41650},
  year={2023}
}

@article{shinn2023reflexion,
  title={Reflexion: Language agents with verbal reinforcement learning},
  author={Shinn, Noah and Cassano, Federico and Gopinath, Ashwin and Narasimhan, Karthik and Yao, Shunyu},
  journal={Advances in Neural Information Processing Systems},
  volume={36},
  pages={8634--8652},
  year={2023}
}

@article{math,
  title={Measuring mathematical problem solving with the math dataset},
  author={Hendrycks, Dan and Burns, Collin and Kadavath, Saurav and Arora, Akul and Basart, Steven and Tang, Eric and Song, Dawn and Steinhardt, Jacob},
  journal={arXiv preprint arXiv:2103.03874},
  year={2021}
}
\bibliographystyle{icml2026}

\appendix
\newpage

\section*{Appendix}

\section{Training details}\label{app:training}
Each model was trained on a single GPU using \texttt{bfloat16} mixed precision for 10000 steps with a batch size of 4 and a max sequence length of 1024 (the typical prompt length depended on the task).
We used the AdamW optimizer with learning rate \texttt{2e-5}, weight decay \texttt{1e-2}, and default hyperparameters otherwise.

\begin{figure*}[!ht]
    \centering
    \includegraphics[width=0.9\linewidth]{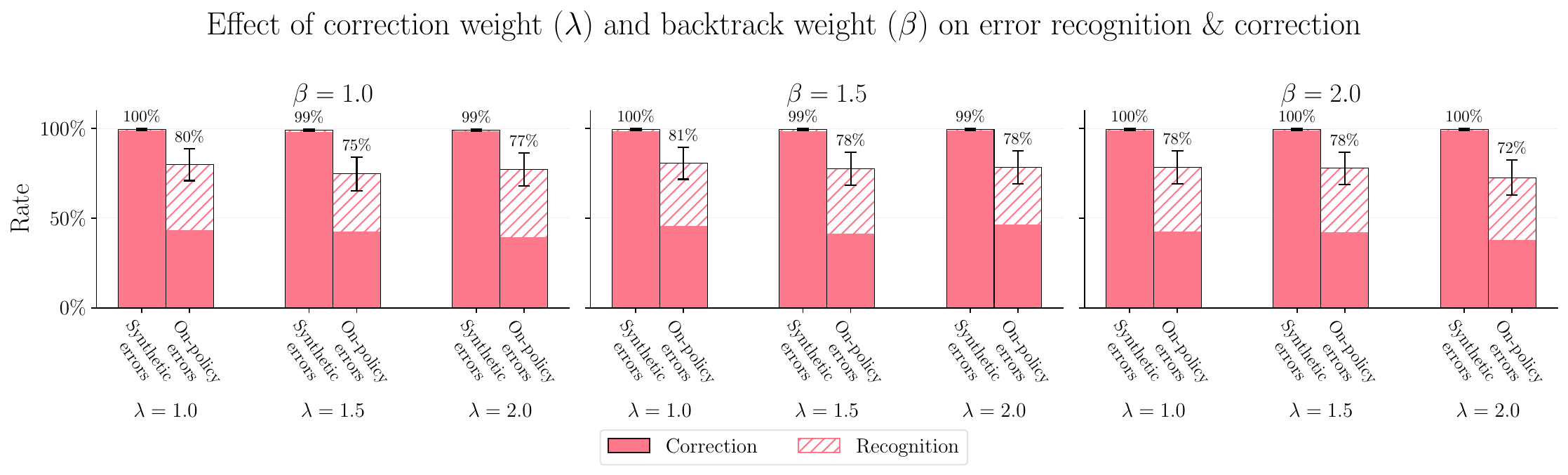}
    \caption{Sweep of correction weight $\lambda$ and backtrack weight $\beta$ across $\{1.0, 1.5, 2.0\} \times \{1.0, 1.5, 2.0\}$ for the tokens in the error correction step for \qwen on multiplication. The error recognition and correction rates remain relatively consistent across the entire sweep.}
    \label{fig:correction_weight}
\end{figure*}

In \Cref{fig:correction_weight} we present our ablation over the token-level weight for the correction and backtrack steps.

\section{CoT examples}\label{app:examples}
\begin{figure*}
    \centering
    \includegraphics[width=0.9\linewidth]{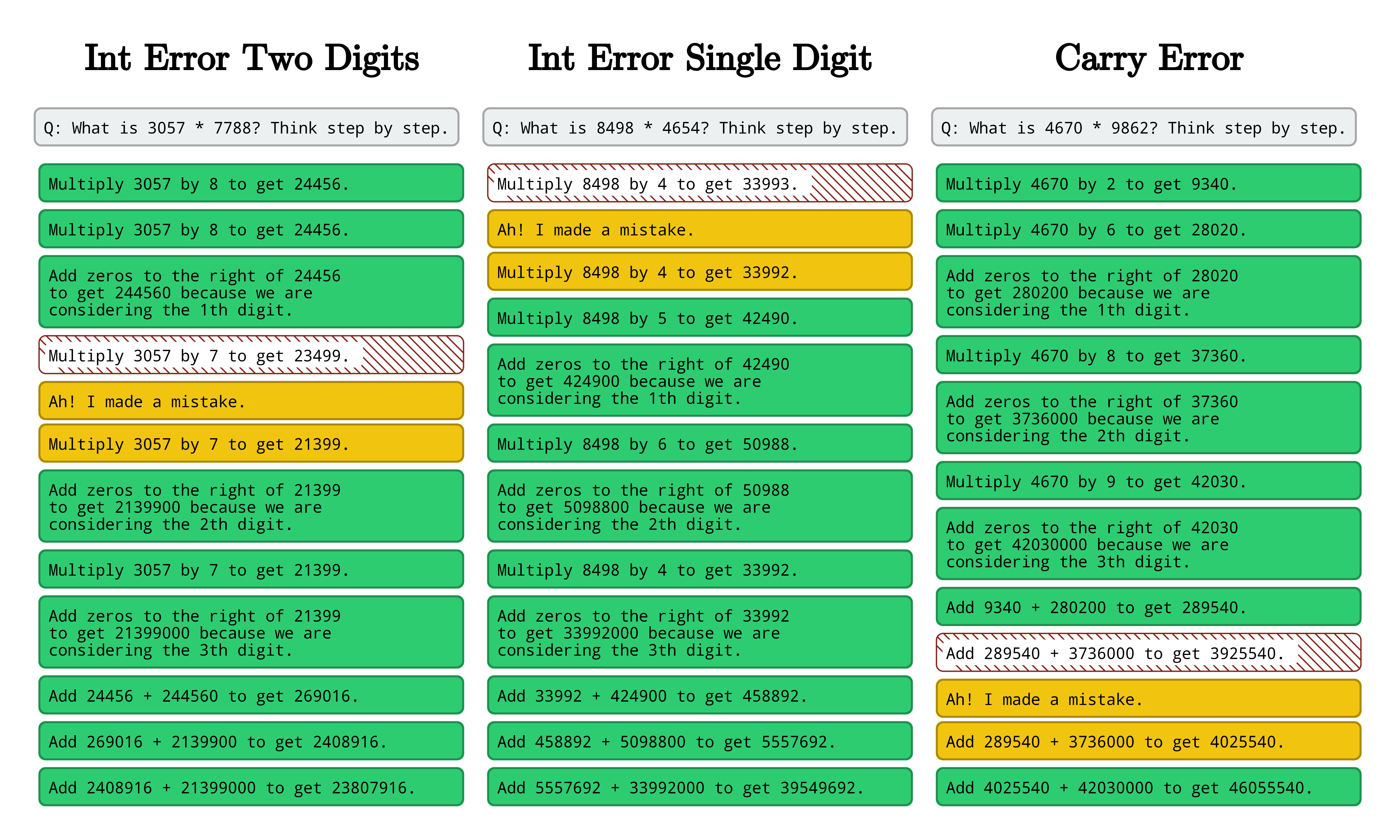}
    \caption{Representative examples of synthetic errors for multiplication. The hatched red steps are the incorrect steps, and the yellow steps are the error recognition/correction steps. See \Cref{subsec:mult-task} for more details on what these types of errors entail.}
    \label{fig:mult_errors_synthetic}
\end{figure*}

\begin{figure*}
    \centering
    \includegraphics[width=0.9\linewidth]{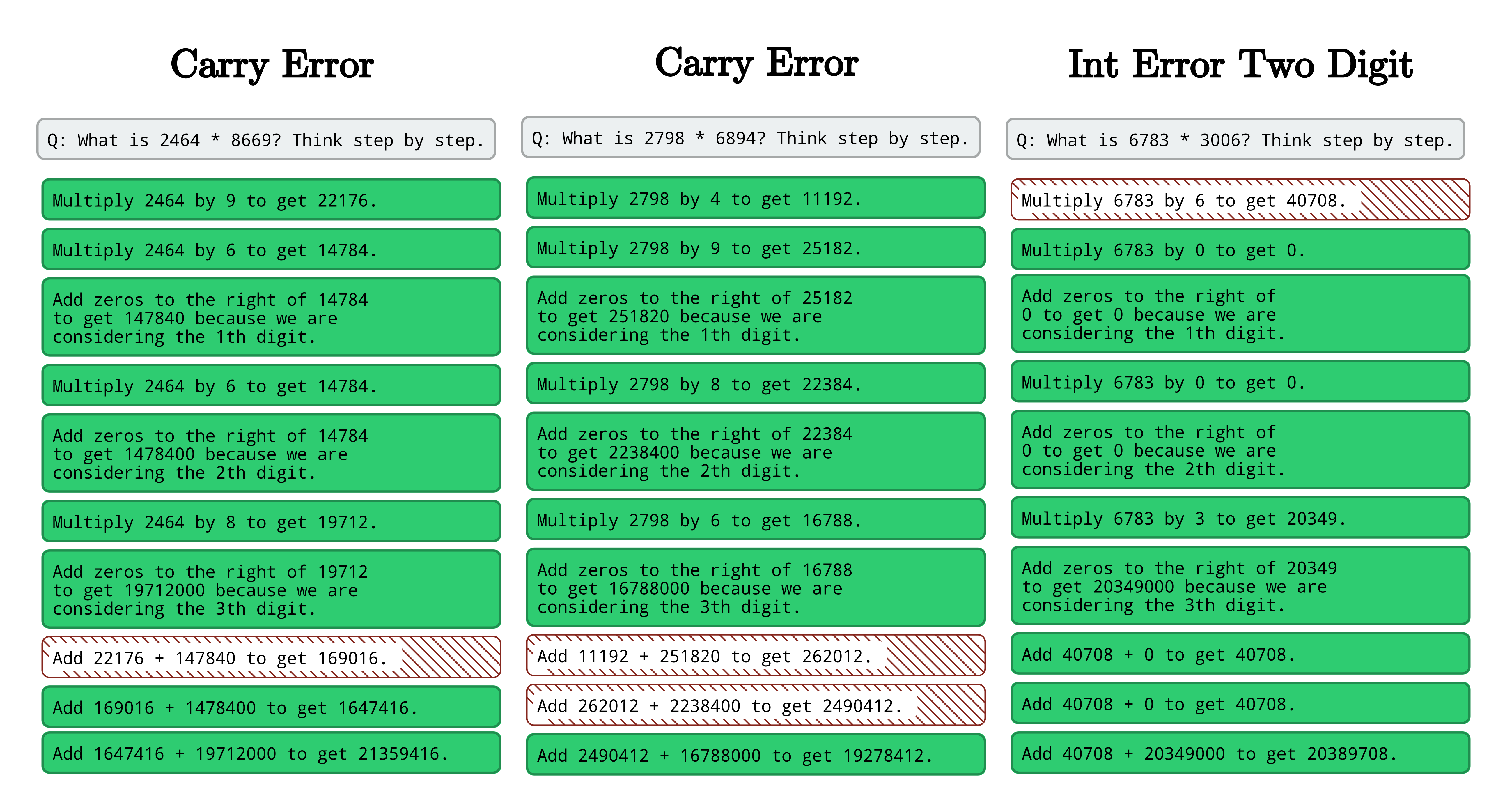}
    \caption{Representative examples of FT errors for multiplication from \qwen. The hatched red steps are the incorrect steps. See \Cref{subsec:mult-task} for more details on what these types of errors entail.}
    \label{fig:mult_errors_ft}
\end{figure*}

\begin{figure*}
    \centering
    \includegraphics[width=0.9\linewidth]{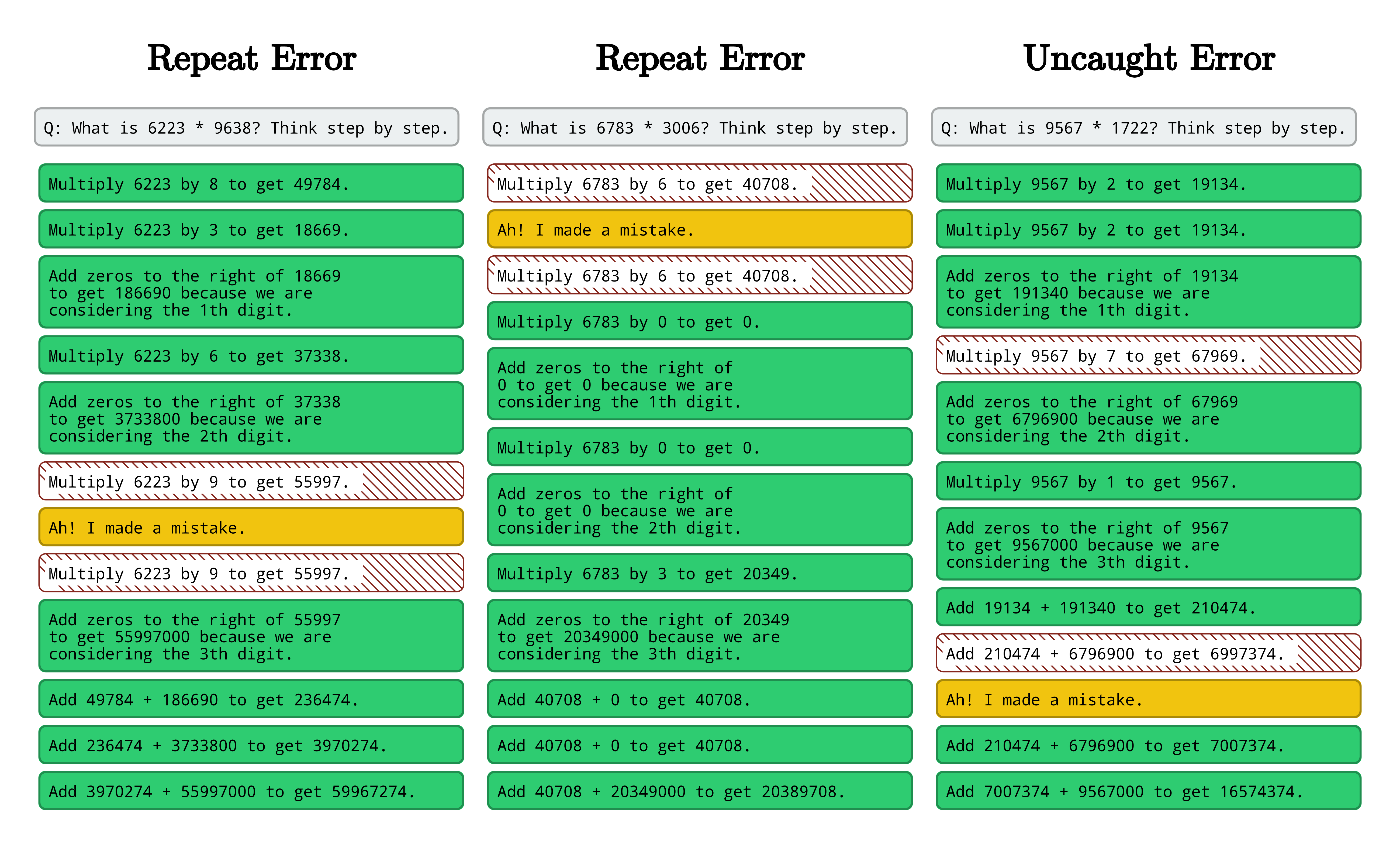}
    \caption{Representative examples of EIFT self-correction failure modes from \qwen. The hatched red steps are incorrect, and the yellow steps are the recognition steps. In the (Left) and (Center) panels we see instances where the EIFT model mimics its original mistake after recognizing it. In the (Right) panel we see a failure to recognize a mistake, along with a successful error correction near the end of the CoT.}
    \label{fig:mult_errors_eift}
\end{figure*}

\begin{figure*}
    \centering
    \includegraphics[width=\linewidth]{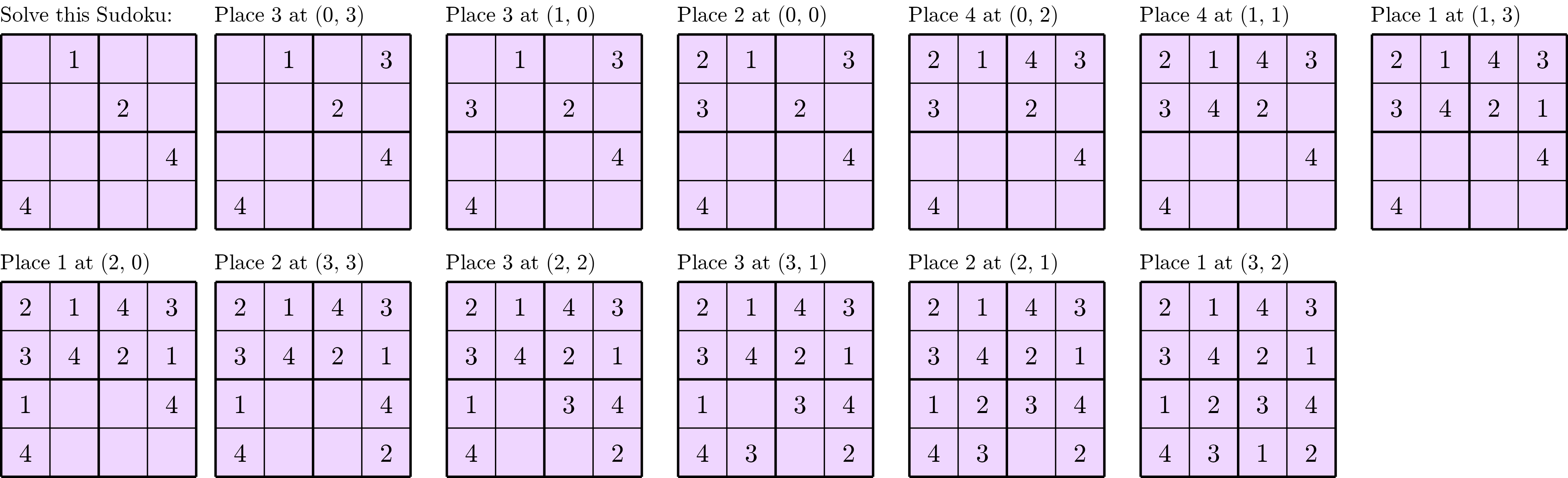}
    \includegraphics[width=\linewidth]{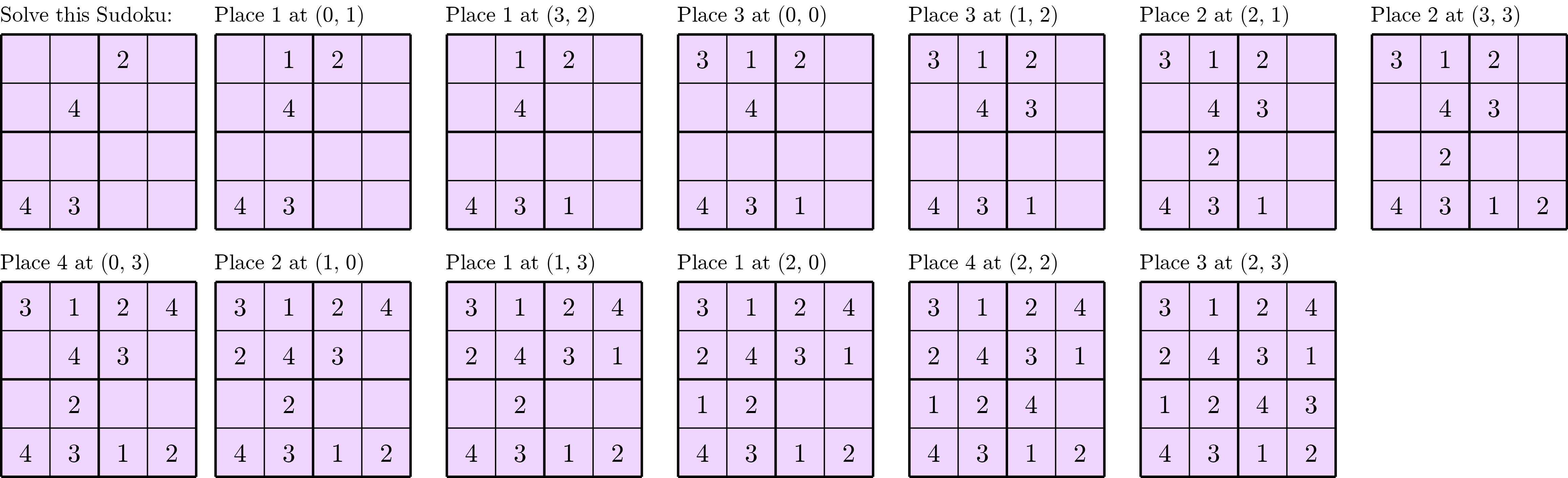}
    \includegraphics[width=\linewidth]{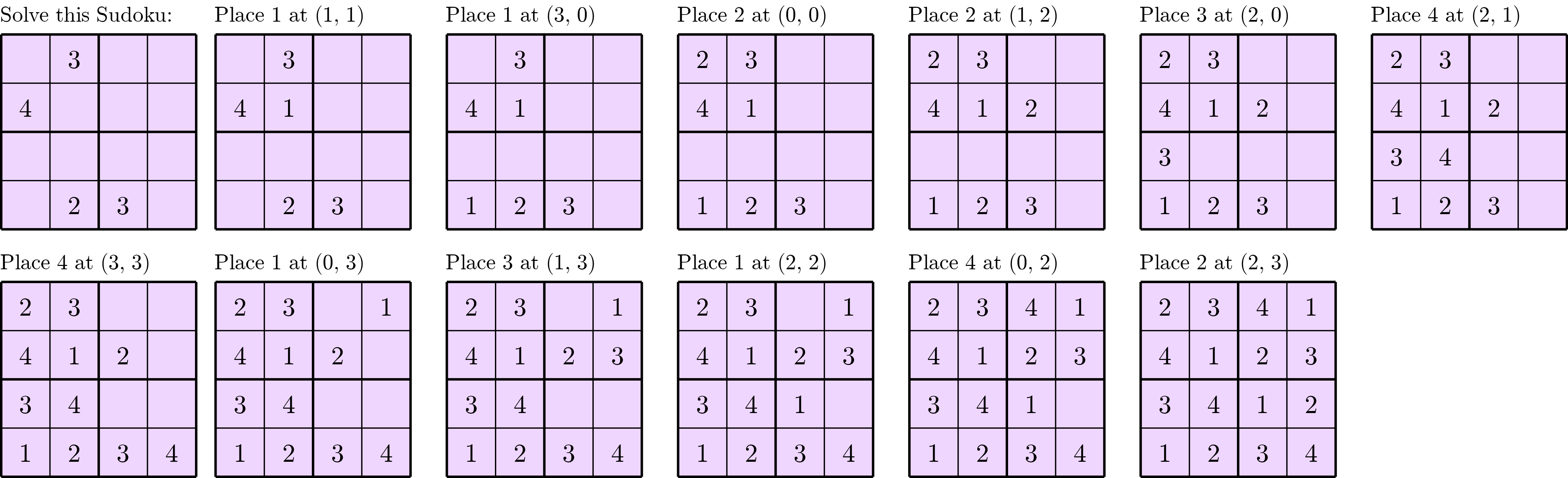}
    \caption{Representative examples of golden chain-of-thoughts (CoTs) from the sudoku task.}
    \label{fig:sudoku_cots}
\end{figure*}

\end{document}